# Large Language Models for Scientific Synthesis, Inference and Explanation

Yizhen Zheng[1]*, Huan Yee Koh[1,3]*, Jiaxin Ju[2]*, Anh T.N. Nguyen[3], Lauren T. May[3,4], Geoffrey I. Webb[1], Shirui Pan[2]

[1]Department of Data Science and Artificial Intelligence, Monash University, Victoria, Australia
[2]School of Information and Communication Technology and Institute for Integrated and Intelligent Systems, Griffith University, Queensland, Australia
[3]Drug Discovery Biology, Monash Institute of Pharmaceutical Sciences, Monash University, Victoria, Australia
[4]Victorian Heart Institute, Monash University, Victoria, Australia
*indicates equal contribution.
indicates corresponding authors: Shirui Pan(s.pan@griffith.edu.au), Geoffrey I. Webb(Geoff.webb@monash.edu), Yizhen Zheng(yizhen.zheng1@monash.edu).

## Abstract

Large language models are a form of artificial intelligence systems whose primary knowledge consists of the statistical patterns, semantic relationships, and syntactical structures of language[1]. Despite their limited forms of 'knowledge,' these systems are adept at numerous complex tasks including creative writing, storytelling, translation, question-answering, summarization, and computer code generation[2,3]. However, they have yet to demonstrate advanced applications in natural science[4,5]. Here we show how large language models can perform scientific synthesis, inference, and explanation. We present a method for using general purpose large language models to make inferences from scientific datasets of the form usually associated with special purpose machine learning algorithms. We show that the large language model can augment this 'knowledge' by synthesizing from the scientific literature. When a conventional machine learning system is augmented with this synthesized and inferred knowledge it can outperform the current state of the art across a range of benchmark tasks for predicting molecular properties. This approach has the further advantage that the large language model can explain the machine learning system's predictions. We anticipate that our framework will open new avenues for AI to accelerate the pace of scientific discovery.

## 1. Introduction

Scientific productivity is in precipitous decline, with the rate of progress in many fields approximately halving every 13 years[6]. As scientific discovery becomes increasingly complex and challenging, traditional methodologies struggle to keep pace, necessitating innovative approaches. Meanwhile, Large Language Model (LLM) Artificial Intelligence systems have shown remarkable capabilities in a wide range of tasks. From creative writing to translating languages, from answering intricate queries[7,8] to code generation[9], their capabilities have been transformative in various domains[1,2,3,10,11]. In this work we show that these LLMs have similar transformational potential in the natural sciences. Particularly, we demonstrate that LLMs can synthesize postulates from the scientific literature, make inferences from scientific data, and elucidate their conclusions with explanations.

LLMs are trained on large corpuses of text, including much of the scientific literature. Notable models like BioBert[12], SciBERT[13], Med-PALM[11], and Galactica[14] are specifically tailored to the scientific domain. Meanwhile, general-purpose LLMs like Falcon[15] integrate extensive scientific literature in their pretraining, including sources such as arXiv and Wikipedia. We demonstrate that these systems have acquired deep abilities to interpret and manipulate the formal scientific language for describing molecules, SMILES strings, along with capability to apply information from the scientific literature in their interpretation. We present a scientific discovery pipeline LLM4SD (Large Language Models for Scientific Discovery) designed to tackle complex molecular property prediction tasks. LLM4SD operates by specifying rules for deriving features from SMILES strings that are relevant to predicting a target feature. Some of these rules are synthesized from the scientific literature that the LLMs encode. Others are inferred from training sets of SMILES strings each labelled with the relevant classes or property values. A standard machine learning model can then be learned from the training data using the rule-based features. Finally, our pipeline utilizes LLMs to produce interpretable outcomes, allowing human experts to ascertain the specific factors influencing the final predictions. We show that this pipeline achieves the current state of the art across 58 benchmark tasks spanning four domains - Physiology, Biophysics, Physical Chemistry and Quantum Mechanics.

Despite these auspicious outcomes, we acknowledge the vastness and intricacy of the scientific discovery landscape; our endeavours have merely scratched the surface. Nonetheless, the strides made by LLM4SD pave the way for deeper exploration, heralding an era where AI-driven insights interweave with human ingenuity to redress the current decline in scientific productivity. Looking ahead, we are optimistic about AI's potential role as a linchpin in the future of scientific discovery, revolutionizing processes and expediting breakthroughs.

## 2. Large Language Models for Scientific Discovery

Our scientific discovery pipeline, LLM4SD, shown in Fig.1 consists of 4 main components: Knowledge Synthesis from the Scientific Literature, Knowledge Inference from Data, Interpretable Model Training and Interpretable Explanation Generation. We demonstrate the application of our pipeline to 58 specialized property prediction tasks across four scientific domains: Physiology, Biophysics, Physical Chemistry, and Quantum Mechanics.

In the Knowledge Synthesis from Literature phase (Fig. 1a), LLMs use pre-trained knowledge from an extensive literature amassed from LLMs' pretraining[14,15] to synthesize domain-specific molecular property prediction rules. Then, in the Knowledge Inference from Data phase (Fig. 1b), LLMs harness their inferential and analytical skills to infer molecular property prediction rules from the patterns in the datasets. These rules can generate features that effectively distinguish between different class instances or predict specific properties, such as a molecule's lipophilicity. This process mirrors how human scientists formulate hypotheses based on observation. In both the knowledge synthesis and inference stages, we require that the rules have either a numerical or categorical measure associated with them. This ensures that the rules can be readily transformed into corresponding functions, which in turn can convert each data instance into a vector of values.

Rules, independently defined by LLMs, transform data instances into vectorized representations, i.e., features. These rule-based features facilitate the training of an interpretable model, e.g., random forest or linear classifier (Fig. 1c). Our preference for training these interpretable models stems from a desire to enhance transparency during predictions. Remarkably, we noted that when enhanced with LLM4SD, traditional interpretable models like random forests can surpass state-of-the-art baselines. These interpretable models, once trained, are adeptly employed for downstream application, encompassing both classification and regression scientific tasks. This entire workflow draws parallels with the methodical approach of human scientists—designing experiments to validate their proposed hypotheses.

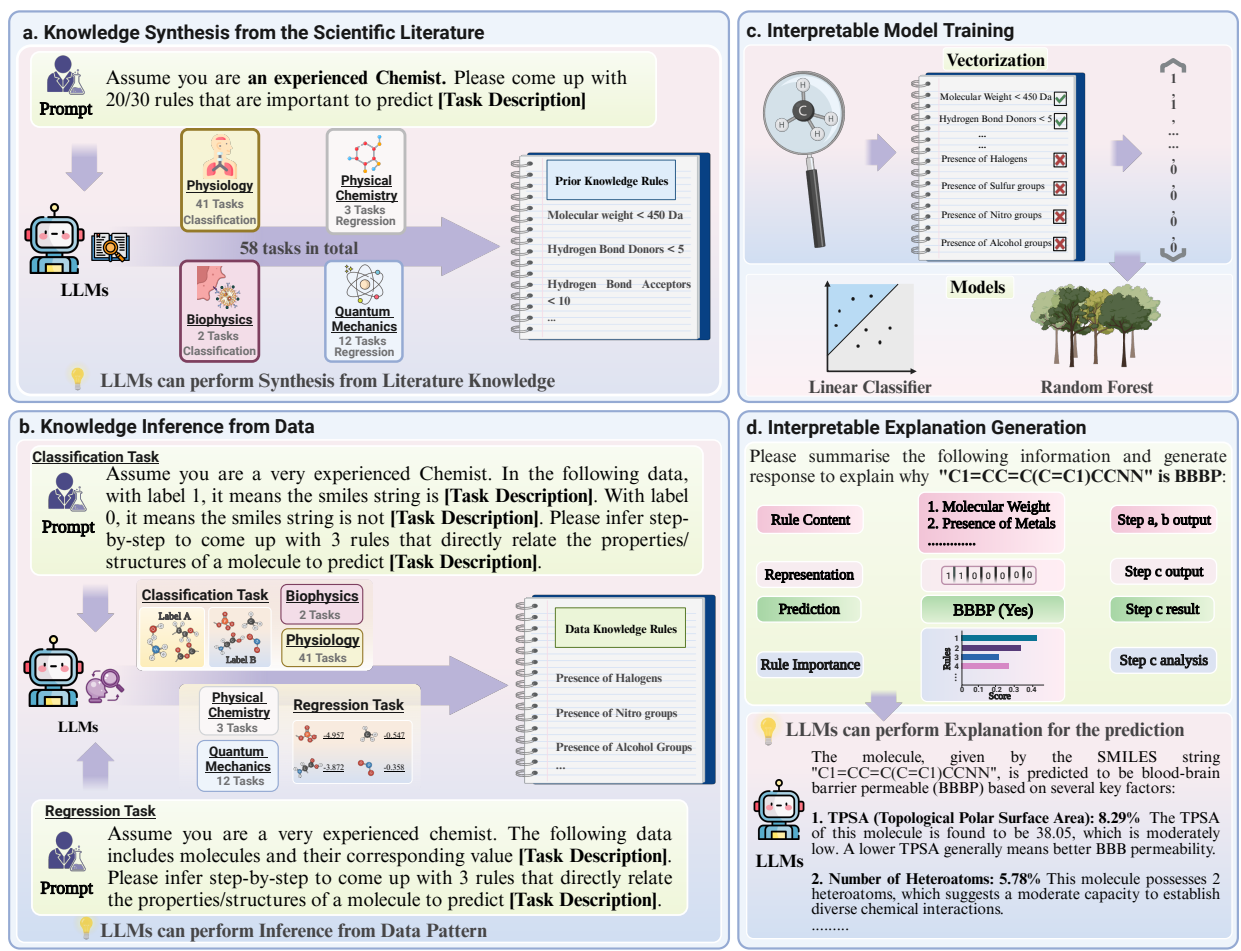

**Fig.1. LLM for Scientific Discovery Pipeline.** (a) Knowledge synthesis from the scientific literature. In this phase, LLMs leverage their pre-trained understanding, amassed from pretraining on a massive body of literature, to synthesize domain-specific rules. (b) Knowledge inference from data. LLMs analyse the intrinsic patterns in the task-specific datasets to identify labelled patterns and infer empirical rules. (c) Interpretable Model Training. Rules formulated from LLMs are harnessed to convert data instances into features, enabling the development of models that are both effective and readily interpretable. (d) Interpretable Explanation Generation. Culminating the process, LLMs assimilate insights from the preceding steps to articulate comprehensive textual explanations, elucidating the rationales behind predictions. In the figure, the percentage value for each factor indicates its importance in concluding the prediction.

In the final stages (Fig. 1d), we tap into the LLMs' adeptness at information summarization. They are tasked to demystify the decision-making mechanism, illuminating how these interpretable models arrive at prediction outcomes based on instance representations, rules, and their respective significance. This clarity and transparency positions LLMs as intuitive partners, enabling scientists to seamlessly interface with and derive insights from the system's decision-making processes. To improve usability for researchers, we have created a web-based application that offers knowledge synthesis, inference, and prediction with explanation functions (see Supplementary Information 3).

By fostering this symbiotic relationship, we not only amplify the efficacy of scientific investigations but also elevate the confidence and trust in AI-assisted conclusions, driving forward the frontier of collaborative research.

## 3. Experiment Results

In this section, we offer a synopsis of LLM4SD's pivotal results spanning the 4 domains of physiology, biophysics, quantum mechanics and physical chemistry. Notably, all results of LLM4SD are obtained based on open-source LLM backbones to ensure reproducibility. Subsequently, we delved into an ablation study of LLM4SD, examining its performance across various LLM backbones[14,15] of differing scales and pretraining datasets.

### 3.1 Overall Performance on Four Domains

To evaluate the versatility of LLM4SD's application, we conducted a comprehensive analysis of its performance across 58 molecular prediction tasks across the 4 domains (Fig.2). Specifically, the physiology domain comprised (Blood-Brain Barrier Penetration) BBBP[16], ClinTox[17], Tox21[18] with 12 tasks, and SIDER[19] with 27 tasks. Biophysics had two tasks, BACE[20] and HIV[18], while physical chemistry had three regression tasks: ESOL[21], FreeSolv[22] and Lipophilicity[18]. Quantum mechanics presented 12 regression tasks under QM9[23]. The detailed description of these tasks is illustrated in the method section (see Methods, 'Datasets'). We compared LL4SD's performance with specialized, state-of-the-art supervised machine learning techniques. These are advanced Graph Neural Networks (GNNs), namely AttrMask[24], GraphCL[25], MolCLR[26], 3DInfomax[27], GraphMVP[28], and MoleBERT[29]. Each model was pre-trained on large datasets with diverse molecular knowledge and then fine-tuned for specific tasks (see Methods). As a standard baseline, we implemented Random Forest[30] with ECFP4[31] as input set features.

Benchmarking LLM4SD against the baseline, LLM4SD demonstrated its superior efficacy and performance (Fig 2). This exemplary performance spanned 58 diverse tasks, from physiology (Extended Data Fig. 1-3) and biophysics (Extended Data Fig. 4) to physical chemistry (Extended Data Fig. 5) and quantum mechanics (Extended Fig. 6).

In both physiology and biophysics, our model outperformed all existing baselines (Fig.2a). Notably, we attained state-of-the-art (SOTA) results in Physiology, raising the AUC-ROC from a previous best of 74.43% to 76.60%, a gain of 2.8%. In Biophysics, our model further enhanced performance, advancing the AUC-ROC from 81.7% to 83.4%, marking a 2.0% improvement. These advancements in physiology and biophysics emphasize the robustness and precision of LLM4SD in tasks that demand intricate biological understanding and modeling.

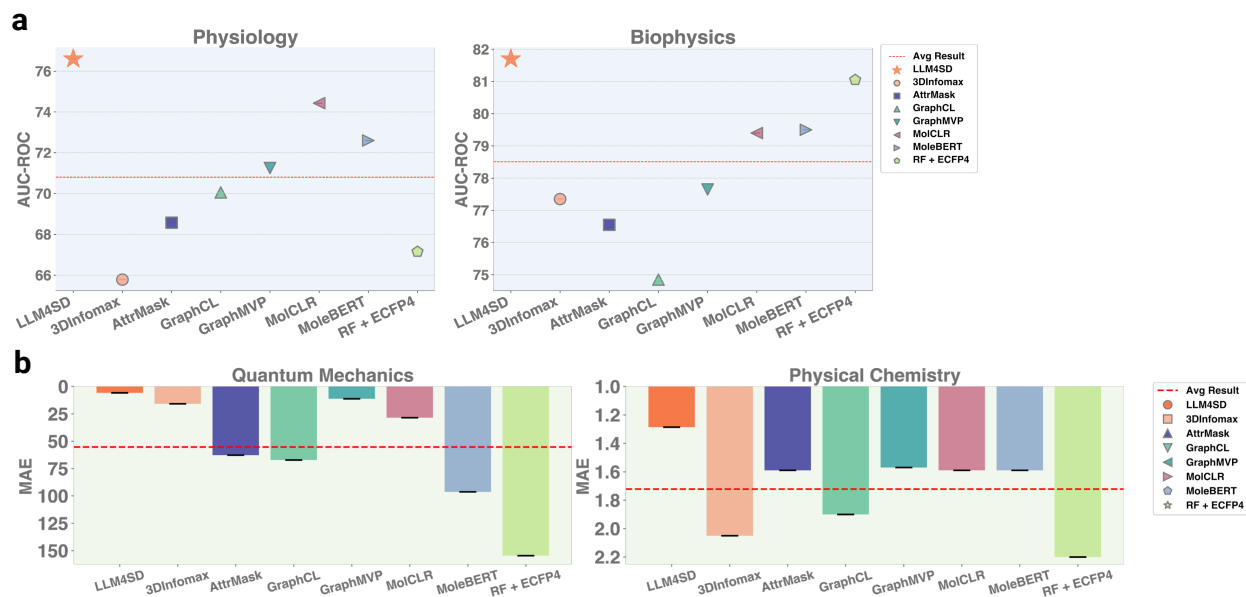

**Fig.2.** Comparison between LLM4SD and baselines across 4 domains. The red dotted line represents the average performance of all baselines. (a) Comparative analysis of model performance versus baselines in physiology and biophysics. (b) Comparative analysis of regression performance: LLM4SD vs. baselines in quantum mechanics and physical chemistry.

On tasks in quantum mechanics and physical chemistry, LLM4SD demonstrated substantial advancements (Fig.2b). In the domain of quantum mechanics, it showed a profound improvement of 48.2% over the best performed baseline, registering an average MAE of 5.8233 across 12 tasks as opposed to 11.2450. Similarly, in physical chemistry, LLM4SD observed a noteworthy enhancement, with the model reaching a MAE of 1.28 marking an 18.5% advancement over the baseline MAE of 1.57. These significant improvements in regression tasks affirm the refined capability of our approach in continuous prediction.

Overall, LLM4SD's marked improvements not only affirm its supremacy over specialized, and often black-box, state-of-the-art models but also highlight its unparalleled ability to synthesize postulates, infer scientific data, and provide insightful explanations. This offers a fresh perspective in computational research and heralds a new direction in scientific endeavors.

### 3.2 Ablation Study

To delve deeper into the intricacies of the LLM4SD pipeline, we conducted an ablation study, focusing on discerning the influence of scale and pretraining datasets on the performance of Large Language Models (LLMs). In addition, we assessed the relative contributions of knowledge synthesis and inference. Our evaluation spanned across a spectrum of foundational LLM backbones, notably the Falcon 7b[15], Falcon 40b[15], Galactica-6.7b[14], and Galactica-30b[14]. Here, we selected open-source LLM backbones to ensure the reproducibility of our work. It is worth noting the distinct differences between the Falcon and Galactica series of LLMs. In particular, the Falcon models are trained for a broad range of applications, imbibing a more general context during their pretraining phase, while the Galactica models are pretrained on mainly scientific literature, making them particularly suitable for science.

### 3.2.1 Effect of Scale

The ablation study of LLM4SD, which compared four open-source LLM backbones, revealed substantial differences among the different LLMs (Fig.3a, b). Particularly within the Falcon series, performance disparities were conspicuous. The Falcon 7b, a smaller model, fell short compared to the Falcon 40b in its range of domain expertise. Notably, it failed to conduct tasks in two key areas: physiology and quantum mechanics, indicating a weaker understanding of scientific challenges and data interpretation.

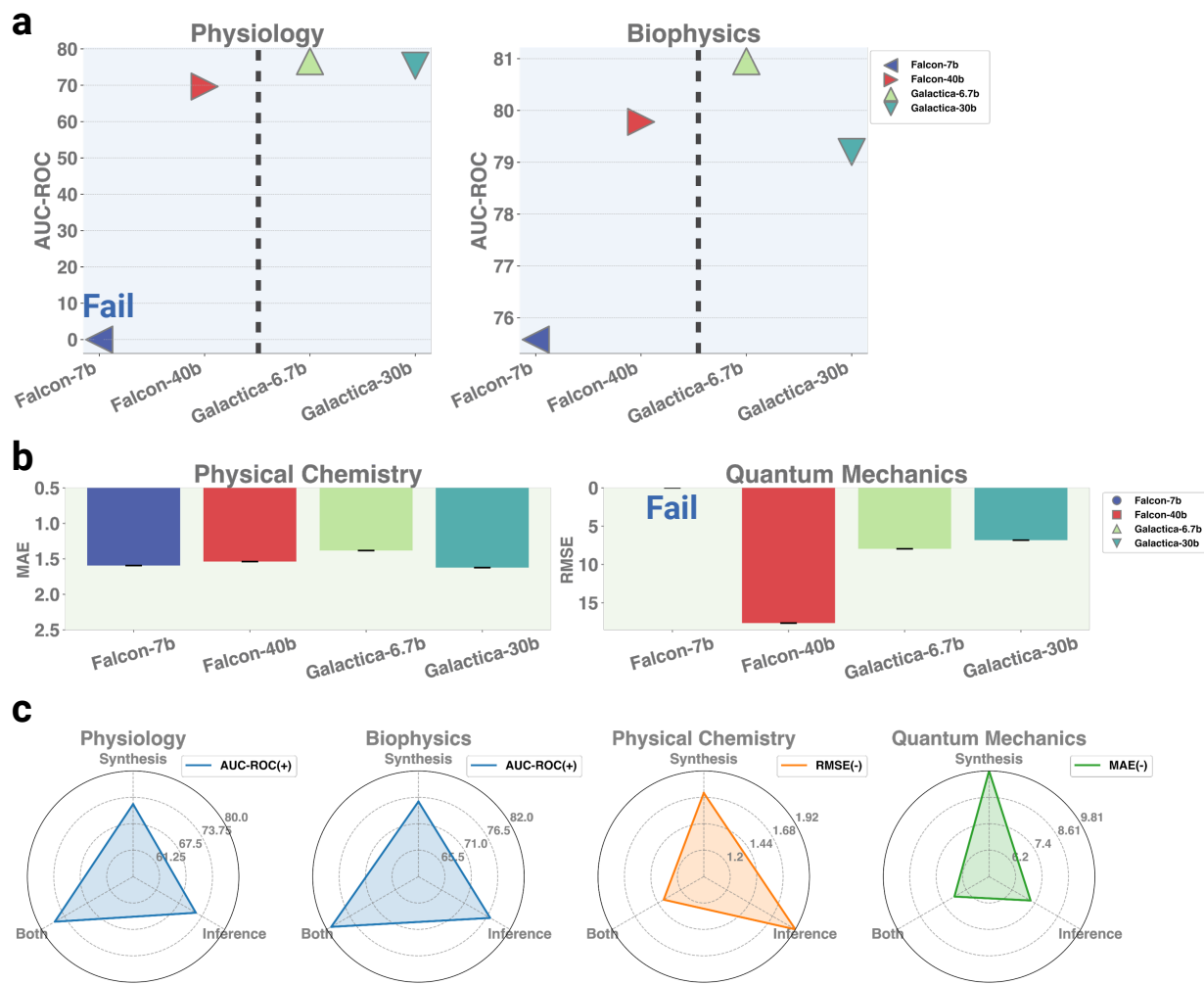

**Fig.3.** Ablation study of LLM4SD. (a) Performance comparison of four open-source LLM backbones for physiology and biophysics. The vertical dashed line in the figure separates the two LLM series: Falcon on the left and Galactica series on the right. (b) Performance comparison of four open-source LLM backbones for physical chemistry and quantum mechanics. (c) Examining the influence of both synthesized and inferred knowledge on the average model performance across four domains. The triangle's color signifies the metric employed for domain-specific tasks. A (+) next to the metric name indicates higher values yield better results, while a (-) suggests the contrary.

Conversely, the Galactica series painted a more nuanced picture. Unlike with the Falcon series, a larger model did not necessarily translate to superior performance. In disciplines such as Physiology, Biophysics, and Physical Chemistry, Galactica 6.7b rivaled the performance of Galactica-30b, despite the latter having more than 4 times the number of parameters. However, in the domain of Quantum Mechanics, the larger Galactica 30b surged ahead, outperforming Galactica 6.7b by a margin of 14%. This variance could be attributed to the intricate and abstract nature of Quantum Mechanics, where the depth and breadth of knowledge encapsulated in the larger model might offer a discernible advantage.

### 3.2.2 Effect of Pretraining Datasets of LLMs

From these observations it becomes evident that an LLM steeped in scientific literature, even if smaller in scale, exhibits a commendable prowess in scientific tasks (Fig.3a, b). Conversely, the Falcon series, designed for general utility, necessitates a more substantial scale to effectively navigate scientific challenges. We postulate that this phenomenon is underpinned by the emergent capabilities[32] inherent to large-scale LLMs. These capabilities empower the more expansive Falcon-40b to bridge the knowledge gap and adapt to scientific tasks. In a broader perspective, despite their relatively modest scale, the Galactica models consistently outperformed the Falcon series, underscoring the pivotal role of domain-specific pretraining.

### 3.2.3 Contributions of knowledge synthesis and inference

In our exploration of LLM4SD with respect to various knowledge sources, we discerned the performance variance arising from the use of rule-based features synthesized from literature, rule-based features inferred from data, and a combined approach. Overall values for these categories were obtained by averaging over results for all tasks in a domain (Fig. 3c).

In a comprehensive assessment of various scientific domains, the combination of synthesis and inference features consistently outperformed individual methods. Specifically, in the field of physiology, an AUC-ROC of 76.38 was achieved using both methods, compared to 72.15 with synthesis alone and 72.12 with inference. Similarly, in biophysics, combining both methods yielded an AUC-ROC of 80.95, surpassing the scores of 75.62 and 77.23 obtained from synthesis and inference features, respectively. In physical chemistry, the combined approach resulted in an RMSE of 1.38, which is notably better than the 1.72 from synthesis features and 1.92 from inference features. Lastly, in Quantum Mechanics, the use of both synthesis and inference features produced a MAE of 6.82, improving upon the values of 9.81 and 7.18 recorded with synthesis and inference alone. Notably, comparing just synthesis with just inference, each outperformed the other in 2 out of the 4 domains.

These observations highlight the value of combining knowledge synthesis from scientific literature with inference from data. Literature imparts foundational theoretical insights, while empirical data identifies further regularities. The fusion of these knowledge facets equips the models with a comprehensive understanding, empowering them to excel across varied tasks and domains.

## 4. Statistical Analysis and Literature Review: Validating Established Rules

With LLM4SD outperforming specialized, state-of-the-art methods, we further validated the rules generated by Galactica-6.7b due to its superior performance and ease of reproducibility. The rules were validated in two ways: statistical tests to confirm the significance of these rules, and literature review to assess whether the rules are discussed in existing scientific literature.

For statistical tests of rules, we employed the Mann-Whitney U test[33] for classification tasks and the linear regression t-test for regression tasks. The Mann-Whitney U test[33] compared the distributions of chosen rule across the two classes of the target variable, thereby evaluating the statistical relevance of the rule's ability to split and distinguish classes. Conversely, the linear regression t-test treated the chosen rule as the independent variable and examined whether its coefficient significantly deviated from 0, reflecting whether the rule contributes to regression prediction.

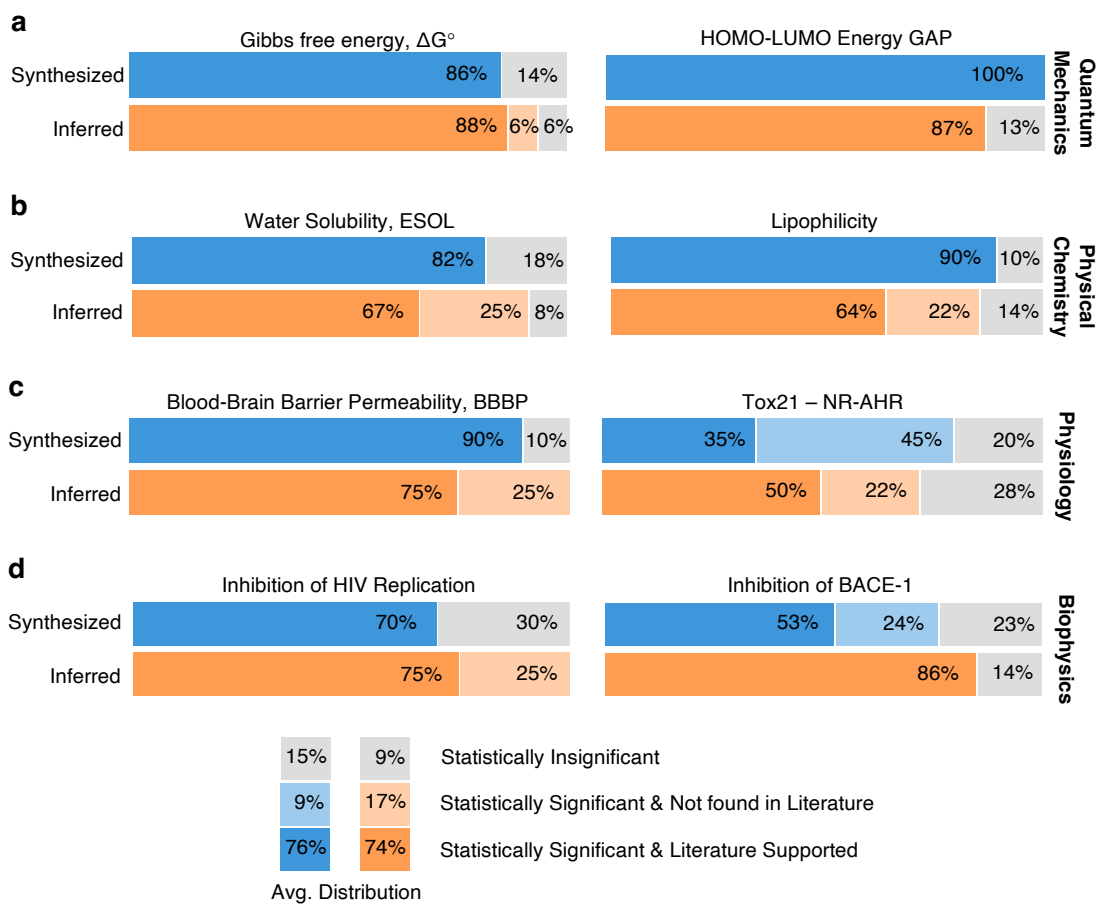

**Fig.4. Literature Review and Statistical Analysis of LLM Rules. a–d.** We conducted statistical analysis and a comprehensive literature review on rules generated by Galactica 6.7b across all four scientific domains, with two tasks evaluated for each domain: (**a**) Physiology, (**b**) Biophysics, (**c**) Quantum Mechanics, and (**d**) Physical Chemistry. In the statistical analysis, the significance of a rule is determined based on the task type: for classification, the Mann-Whitney U test[33] compares the difference in distributions of chosen rule across the two classes of the target variable; and for regression, the linear regression t-test[26] treats the chosen rule as the independent variable and examined whether its coefficient significantly deviated from 0, reflecting whether the rule contributes to prediction. In both cases, we used a 0.05 p-value threshold to determine rule significance. In the literature review, we assessed the prevalence of a rule in existing literature. With statistical analysis and literature review, each rule is categorized into one of three classes: statistically significant and literature supported; statistically significant and not found in literature; and statistically insignificant. Across all tasks, literature-synthesized knowledge rules were generally both prevalent in existing literature and statistically significant. In contrast, empirically inferred data rules yielded mixed results, with some easily found in existing literature and others not identified by the researchers.

We further carried out a comprehensive review with in-domain experts to evaluate the prevalence of a rule in existing literature (see Supplementary Information 4: Literature Review(Example)). After cross-referencing the rules with scientific literature, we categorized each rule into one of three classes: statistically significant and literature supported; statistically significant and not found in literature; and statistically insignificant (Fig. 4).

## 4.1 Knowledge Synthesis from Scientific Literature

We discovered that most of the synthesized rules we examined are readily available in existing scholarly works. Notably, an overwhelming majority (85%) of these rules were statistically significant in indicating the target labels across all selected tasks, affirming our pipeline's ability to summarize rules from scientific literature that were the most important for different tasks and domains.

Importantly, except for BACE[20] and Tox21-NR-Ahr[18], we found no instances where statistically significant rules were absent from existing literature (Fig. 4). This aligns with the design of our pipeline: without analyzing the data, LLMs tend to aggregate and summarize existing knowledge. To illustrate, in the context of BBBP, the rules generated by our pipeline were consistent with well-established determinants such as molecular weight, lipophilicity, distribution coefficient, topological polar surface area, and hydrogen bonds[34-36]. These findings validate the robustness and reliability of our pipeline in leveraging LLMs to summarize existing scientific literature.

## 4.2 Knowledge Inference from Data

We found that an average of 91.3% of the inferred rules were statistically significant, higher than synthesized rules (Fig. 4). Of these, an average of 74% rules were already documented in existing scientific literature, while 17.3% were not identified by researchers. These latter rules were primarily associated with data patterns that are not widely discussed in the literature but can be inferred from our task dataset, such as obscure molecular substructures that influence target labels like the Gibbs free energy ($\Delta G°$) of a molecule.

In contrast, to the knowledge synthesized from literature, we found that 6 out of 8 tasks have statistically significant rules that were absent from existing literature. This suggests that the rules produced by LLM4SD are not merely a result of the LLM's textual memorization during pre-training. Instead, the inferred rules reflect a genuine capability to derive meaningful rules from data based on the specific task.

Our case studies further substantiated the utility of these unidentified but significant rules. For instance, in BBBP where 38% of rules are significant but unidentified, Galactica 6.7B pinpointed the carbonyl functional group and fragment rings as key determinants of a molecule's BBBP. We hypothesize that these features are crucial for calculating a molecule's cross-sectional area, which in turn influences its orientation in lipid-water interfaces—factors vital for membrane partitioning and permeation[37]. Intriguingly, this suggests that our pipeline enables LLMs to infer what we term as second-order features. These are features that may not be immediately obvious or widely recognized but are consistent with established scientific principles in literature. In doing so, LLMs not only corroborate existing knowledge but also apply existing knowledge in interpreting data, thereby enriching the current scientific discourse.

By leveraging LLMs, our pipeline not only validates well-established scientific principles but also uncovers less documented and even potentially novel rules. This facilitates a more effective and transparent interaction between scientists and the AI system, enhancing both the quality and trustworthiness of the research output. Moreover, the statistically robust but underrepresented rules we identified could serve as promising avenues for future scientific exploration, thereby advancing the frontier of collaborative, AI-assisted research.

## 5. Discussion

In our exploration, we unveil the capabilities of LLM4SD through our specially designed pipeline, enabling LLMs to excel in scientific synthesis, inference, and explanation. Through seamless integration with our proposed architecture, LLMs exhibit state-of-the-art (SOTA) performance across a vast expanse of four domains. The inherent versatility of LLM4SD stands as a testament to its potential, making it poised for broader applications across varied domains, thus magnifying its relevance in the current scientific landscape.

Scientific discovery, vast in scope, is constantly evolving with our expanding understanding of the universe. Our study, ambitious in its intent, captures 58 tasks across four distinct domains, providing a glimpse into the immense reservoir of scientific knowledge. While this study serves as a pioneering beacon, demonstrating LLMs' transformative capabilities, it also signals the beginning of a broader exploration. We envision further expansion, integrating more diverse tasks and domains, pushing LLMs to their full potential and reshaping the boundaries of scientific inquiry.

Harnessing the immense power of AI-driven models for scientific discovery brings along its ethical challenges. The vast capabilities of such models, while revolutionizing our understanding, also raise concerns of potential misuse, especially in sensitive domains like biophysics and quantum mechanics. The reliance on machine synthesis and interpretation might overshadow the indispensable human element of scrutiny and ethics in research. As we plunge deeper into the AI era, it's crucial to tread with caution, balancing advancements with rigorous oversight and an unwavering commitment to ethical rigor.

As we gaze towards the horizon, the potential trajectory for LLM4SD is compelling. We anticipate a future where the nexus between LLMs and advanced scientific toolkits deepens. As computational capabilities grow and scientific knowledge expands, our pipeline stands poised for evolutionary enhancements. Our steadfast goal is to harmoniously fuse artificial intelligence with myriad scientific arenas, unlocking novel insights and pioneering avenues previously unimagined.

# References

1. Frank MC. Baby steps in evaluating the capacities of large language models. *Nat Rev Psychol.* (2023).

2. Brown T, Mann B, Ryder N, Subbiah M, Kaplan JD, Dhariwal P, Neelakantan A, Shyam P, Sastry G, Askell A, Agarwal S. Language models are few-shot learners. *Adv Neural Inf Process Syst.* (2020); 33:1877-901.

3. OpenAI. GPT-4 Technical Report. *preprint.* (2023);

4. Birhane A, Kasirzadeh A, Leslie D et al. Science in the age of large language models. *Nat Rev Phys.* (2023); 5:277-280.

5. Gilbert S, Harvey H, Melvin T et al. Large language model AI chatbots require approval as medical devices. *Nat Med.* (2023).

6. Bloom, Nicholas, Charles I. Jones, John Van Reenen, and Michael Webb. "Are ideas getting harder to find?." *American Economic Review* 110, no. 4 (2020): 1104-1144.

7. Wei, Jason, Xuezhi Wang, Dale Schuurmans, Maarten Bosma, Fei Xia, Ed Chi, Quoc V. Le, and Denny Zhou. "Chain-of-thought prompting elicits reasoning in large language models." *Advances in Neural Information Processing Systems* 35 (2022): 24824-24837.

8. Zhou, Denny, Nathanael Schärli, Le Hou, Jason Wei, Nathan Scales, Xuezhi Wang, Dale Schuurmans et al. "Least-to-most prompting enables complex reasoning in large language models." *ICLR* (2023).

9. Rozière, Baptiste, Jonas Gehring, Fabian Gloeckle, Sten Sootla, Itai Gat, Xiaoqing Ell en Tan, Yossi Adi et al. "Code Llama: Open Foundation Models for Code." *arXiv preprint arXiv:2308.12950* (2023).

10. Jiang, Lavender Yao, Xujin Chris Liu, Nima Pour Nejatian, Mustafa Nasir-Moin, Duo Wang, Anas Abidin, Kevin Eaton et al. "Health system-scale language models are all-purpose prediction engines." *Nature* (2023): 1-6.

11. Singhal, Karan, Shekoofeh Azizi, Tao Tu, S. Sara Mahdavi, Jason Wei, Hyung Won Chung, Nathan Scales et al. "Large language models encode clinical knowledge." *Nature* (2023).

12. Lee, Jinhyuk, Wonjin Yoon, Sungdong Kim, Donghyeon Kim, Sunkyu Kim, Chan Ho So, and Jaewoo Kang. "BioBERT: a pre-trained biomedical language representation model for biomedical text mining." *Bioinformatics* 36, no. 4 (2020): 1234-1240.

13. Beltagy, Iz, Kyle Lo, and Arman Cohan. "SciBERT: A pretrained language model for scientific text." *ACL* (2019).

14. Taylor, Ross, Marcin Kardas, Guillem Cucurull, Thomas Scialom, Anthony Hartshorn, Elvis Saravia, Andrew Poulton, Viktor Kerkez, and Robert Stojnic. "Galactica: A large language model for science." *Preprint* (2022).

15. Almazrouei, Ebtesam, Hamza Alobeidli, Abdulaziz Alshamsi, Alessandro Cappelli, Ruxandra Cojocaru, Merouane Debbah, Etienne Goffinet et al. "Falcon-40B: an open large language model with state-of-the-art performance." Technical report, Technology Innovation Institute; (2023).

16. Martins, Ines Filipa, Ana L. Teixeira, Luis Pinheiro, and Andre O. Falcao. "A Bayesian approach to in silico blood-brain barrier penetration modeling." *Journal of Chemical Information and Modeling* 52, no. 6 (2012): 1686-1697.

17. Gayvert, Kaitlyn M., Neel S. Madhukar, and Olivier Elemento. "A data-driven approach to predicting successes and failures of clinical trials." *Cell Chemical Biology* 23, no. 10 (2016): 1294-1301.

18. Wu, Z., Ramsundar, B., Feinberg, E. N., Gomes, J., Geniesse, C., Pappu, A. S., ... & Pande, V. (2018). MoleculeNet: a benchmark for molecular machine learning. *Chemical Science*, *9*(2), 513-530.

19. Kuhn, Michael, Ivica Letunic, Lars Juhl Jensen, and Peer Bork. "The SIDER database of drugs and side effects." *Nucleic Acids Research* 44, no. D1 (2016): D1075-D1079.

20. Subramanian, Govindan, Bharath Ramsundar, Vijay Pande, and Rajiah Aldrin Denny. "Computational modeling of β-secretase 1 (BACE-1) inhibitors using ligand based approaches." *Journal of Chemical Information and Modeling 56*, no. 10 (2016): 1936-1949.25

21. Delaney, John S. "ESOL: estimating aqueous solubility directly from molecular structure." *Journal of Chemical Information and Computer Sciences 44*, no. 3 (2004): 1000-1005.

22. Mobley, David L., and J. Peter Guthrie. "FreeSolv: a database of experimental and calculated hydration free energies, with input files." *Journal of Computer-Aided Molecular Design 28* (2014): 711-720.

23. Ramakrishnan, Raghunathan, Pavlo O. Dral, Matthias Rupp, and O. Anatole Von Lilienfeld. "Quantum chemistry structures and properties of 134 kilo molecules." *Scientific Data* 1, no. 1 (2014): 1-7.

24. Hu, Weihua, Bowen Liu, Joseph Gomes, Marinka Zitnik, Percy Liang, Vijay Pande, and Jure Leskovec. "Strategies for Pre-training Graph Neural Networks." In International Conference on Learning Representations. 2019.

25. You, Yuning, Tianlong Chen, Yongduo Sui, Ting Chen, Zhangyang Wang, and Yang Shen. "Graph contrastive learning with augmentations." Advances in Neural Information Processing Systems 33 (2020): 5812-5823.

26. Wang, Yuyang, Jianren Wang, Zhonglin Cao, and Amir Barati Farimani. "Molecular contrastive learning of representations via graph neural networks." Nature Machine Intelligence 4, no. 3 (2022): 279-287.

27. Stärk, Hannes, Dominique Beaini, Gabriele Corso, Prudencio Tossou, Christian Dallago, Stephan Günnemann, and Pietro Liò. "3d infomax improves gnns for molecular property prediction." In International Conference on Machine Learning, pp. 20479-20502. PMLR, 2022.

28. Liu, Shengchao, Hanchen Wang, Weiyang Liu, Joan Lasenby, Hongyu Guo, and Jian Tang. "Pre-training Molecular Graph Representation with 3D Geometry." *In International Conference on Learning Representations*. 2021.

29. Xia, Jun, Chengshuai Zhao, Bozhen Hu, Zhangyang Gao, Cheng Tan, Yue Liu, Siyuan Li, and Stan Z. Li. "Mole-bert: Rethinking pre-training graph neural networks for molecules." In *The Eleventh International Conference on Learning Representations*. 2022.

30. Breiman, Leo. "Random forests." *Machine Learning* 45 (2001): 5-32.

31. Rogers, David, and Mathew Hahn. "Extended-connectivity fingerprints." *Journal of Chemical Information and Modeling* 50, no. 5 (2010): 742-754.

32. Wei, Jason, Yi Tay, Rishi Bommasani, Colin Raffel, Barret Zoph, Sebastian Borgeaud, Dani Yogatama et al. "Emergent abilities of large language models." *TMLR* (2022).

33. McKnight, Patrick E., and Julius Najab. "Mann-Whitney U Test." *The Corsini Encyclopedia of Psychology* (2010): 1-1.

34. Wager, T. T. et al. Defining desirable central nervous system drug space through the alignment of molecular properties, in vitro adme, and safety attributes. *ACS Chemical Neuroscience* 1, 420–434 (2010)

35. Wager, T. T., Hou, X., Verhoest, P. R. & Villalobos, A. Moving beyond rules: the development of a central nervous system multiparameter optimization (cns mpo) approach to enable alignment of druglike properties. *ACS Chemical Neuroscience* 1, 435–449 (2010).

36. Geldenhuys, W. J., Mohammad, A. S., Adkins, C. E. & Lockman, P. R. Molecular determinants of blood–brain barrier permeation. *Therapetuic. Delivery* 6, 961–971 (2015).

37. Gerebtzoff, G. & Seelig, A. In silico prediction of blood- brain barrier permeation using the calculated molecular cross-sectional area as main parameter. J. *Chemical Information Modeling* 46, 2638–2650 (2006). 14/14.

## Methods

### Datasets:

We conducted a thorough evaluation of LLM4SD, covering 58 subtasks across four unique domains for a robust assessment. The physiology domain included 41 tasks like BBBP, ClinTox, and the 12-task Tox21, ranging from NR-AR to SR-p53, along with the 27-task SIDER suite covering various medical conditions. Biophysics offered 2 classification tasks: BACE and HIV. In physical chemistry, we addressed three regression tasks: ESOL, FreeSolv, and Lipophilicity, while the Quantum mechanics domain presented 12 regression tasks within the QM9 dataset, exploring properties from mu to G, providing a comprehensive insight into LLM4SD's capabilities.

### Physiology.

**BBBP:** The BBBP dataset contains 2,039 instances, each representing unique compounds labeled based on their permeability properties. Predicting which molecules can cross this barrier is paramount for drug development, especially for neurological conditions.

**ClinTox:** The ClinTox dataset, with 1,478 instances, provides comprehensive information on the toxicological properties of various compounds.

**Tox21:** With 7,831 instances, the Tox21 dataset is a collaborative effort to identify environmental toxicants. Its 12 classification tasks focus on specific biological targets or pathways. The Nuclear Receptor (NR) tasks, namely NR-AhR, NR-AR, NR-AR-LBD, NR-Aromatase, NR-ER, NR-ER-LBD, and NR-PPAR-gamma, examine interactions with intracellular proteins influencing gene expression and potential toxic effects. The Stress Response (SR) tasks, including SR-ARE, SR-ATAD5, SR-HSE, SR-MMP, and SR-p53, explore the impact of chemicals on stress-related cellular pathways.

**SIDER:** The SIDER dataset, with 1,427 instances, offers detailed data on medication side effects. Each task in this dataset relates to a specific adverse drug reaction, aiding researchers in understanding and predicting potential drug side effects. All **27 classification tasks** are: 1) Hepatobiliary disorders 2) Metabolism and nutrition disorders 3) Product issues 4) Eye disorders 5) Investigations 6) Musculoskeletal and connective tissue disorders 7) Gastrointestinal disorders 8) Social circumstances 9) Immune system disorders 10) Reproductive system and breast disorders 11) Neoplasms benign, malignant and unspecified (incl cysts and polyps). 12) General disorders and administration site conditions 13) Endocrine disorders 14) Surgical and medical procedures 15) Vascular disorders 16) Blood and lymphatic system disorders 17) Skin and subcutaneous tissue disorders 18) Congenital, familial and genetic disorders 19) Infections and infestations 20) Respiratory, thoracic and mediastinal disorders 21) Psychiatric disorders 22) Renal and urinary disorders 23) Pregnancy, puerperium and perinatal conditions 24) Ear and labyrinth disorders 25) Cardiac disorders 26) Nervous system disorders 27) Injury, poisoning and procedural complications.

### Biophysics.

**HIV:** With a collection of 17,930 instances, the HIV dataset offers a comprehensive repository of molecules, represented in the SMILES format. This dataset is instrumental in the classification of compounds based on their potential inhibitory effects against HIV.

**BACE:** The BACE dataset, comprising 11,908 instances, is a curated collection of molecules, each represented in the SMILES format. This dataset is tailored for classification tasks, aiming to discern molecules that can inhibit the BACE-1 enzyme. By analyzing the molecules within this dataset, researchers can glean insights into the structural features that confer inhibitory properties against BACE-1.

**Physical Chemistry.**

**ESOL:** The ESOL dataset, comprising 1,128 instances, is a curated collection that delves into the solubility of molecules in water. By analyzing the ESOL dataset, researchers can gain a deeper understanding of the molecular features that dictate solubility, thereby aiding in the design of compounds with optimal solubility profiles. Each entry in this dataset is represented using the SMILES notation, a universal language for describing the structure of chemical species.

**FreeSolv:** With 642 instances, the FreeSolv dataset provides comprehensive data on the hydration free energy of molecules. This dataset is pivotal for researchers aiming to predict how molecules interact with water, which has implications for drug solubility and stability. Each molecule in the FreeSolv dataset is also represented using the SMILES notation.

**Lipophilicity:** Lipophilicity is a fundamental property that influences the absorption, distribution, metabolism, and excretion of drugs. The Lipophilicity dataset with 4200 compounds offers a rich resource for understanding this property. Analyzing this dataset allows researchers to discern the molecular attributes that contribute to a compound's lipophilicity, guiding the synthesis of molecules with desired pharmacokinetic properties. Like the other datasets in this domain, each entry is denoted using the SMILES notation.

**Quantum Mechanics.**

**QM9:** The Quantum Mechanics domain, central to understanding the fundamental properties of matter, is exemplified in our evaluation through the QM9 dataset. Comprising 133,885 instances, the QM9 dataset provides a comprehensive exploration of molecules' quantum mechanical attributes, essential for diverse applications from material science to pharmaceuticals. It includes 12 tasks: $\mu$ (Dipole Moment), $\alpha$ (Polarizability), $R^2$ (Squared Radius), ZPVE (Zero-Point Vibrational Energy), $C_v$ (Heat Capacity at Constant Volume), $\Delta_\epsilon$ (Energy Gap), $\epsilon_{HOMO}$ (Highest Occupied Molecular Orbital Energy), $\epsilon_{LUMO}$ (Lowest Unoccupied Molecular Orbital Energy), $U_0$ (Internal Energy at 0 Kelvin), U (Internal Energy at Standard State), H (Enthalpy), G (Gibbs Free Energy).

## Baselines:

We rigorously assessed our pipeline in comparison to specialized, state-of-the-art supervised machine learning methods. For conventional approaches, we employed Random Forest[30], using ECFP4[31] as the input feature set. We also considered state-of-the-art Graph Neural Networks (GNNs), including Attribute Masking (AttrMask)[24], GraphCL[25], MolCLR[26], 3DInfomax[27], GraphMVP[28], and MoleBERT[29]. Each of these models was initialized with pre-trained weights and subsequently fine-tuned for specific tasks.

In summary, AttrMask pre-training involves teaching the GNN to predict randomly masked atom and bond attributes within molecular graphs. GraphCL and MolCLR use graph augmentations for a contrastive learning objective, aimed at maximizing the similarity between augmentations originating from the same molecule while minimizing similarity between augmentations from different molecules. GraphMVP and 3DInfomax leverage existing 3D molecular datasets to pre-train models capable of deducing 3D molecular geometry from 2D graphs, by optimizing mutual information between 3D summary vectors and GNN graph representations. Finally, MoleBERT, the recent state-of-the-art method, employs a VQ-VAE-based tokenizer to diversify atom vocabulary, thereby balancing dominant and rare atoms. It uses Masked Atoms Modeling (MAM) and Triplet Masked Contrastive Learning (TMCL) for node and graph-level pre-training, respectively.

## LLM for Scientific Discovery Pipeline

In this section, we detail the proposed pipeline and the techniques used to align them with the requirements of molecular property prediction tasks. Instead of merely prompting LLMs to generate scientific hypotheses[38] or training them for direct predictions[39], LLM4SD emulates how human experts conduct scientific research. This includes synthesizing knowledge from literature, inferring hypotheses from datasets, validating findings through experiments, and elucidating the rationale behind predictions.

### Knowledge Synthesis from the Scientific Literature

LLMs are usually pretrained on large corpora of text data that include books, articles, websites and other written content. This extensive pretraining helps LLMs to learn the structure of the language, recognize patterns, understand context and acquire a wide-ranging knowledge of facts and concepts. Thus, the goal of the knowledge synthesis process is to extract relevant features from the vast pool of the knowledge that a LLM possesses from the pretraining stage.

To achieve this, we first instruct the LLM to adopt the persona of an experienced chemist, and then engage it to identify pertinent features based on its existing knowledge. This form of role-playing prompt facilitates the knowledge mining process to mimic how human experts solve real-world challenges. For example, when a chemist needs to predict the bioactivity or BBBP of a molecule, they often apply feature-related rules such as number of hydrogen bond donors/acceptors, molecular weight, and logP. We require that the features identified by LLMs can be measured with a numerical or categorical measure to enable their transcription into corresponding functions.

### Knowledge Inference from Data

The objective of knowledge inference form data is to harness the powerful reasoning skills of LLMs to identify relevant features by analyzing the given data. Given their impressive ability to solve mathematical problems and identify patterns, we conjecture that LLMs have the capacity to discern common patterns within groups of molecules based on its scientific understanding. To validate this hypothesis, we provide LLMs with an instruction and several batches of sampled instances with their corresponding class labels or instance property values. In the instruction, the LLM is tasked with analyzing patterns from provided data to identify features that effectively discriminate between two classes of instances or predict their property values. As a result, LLMs will come up with rules distilled from the analysis for each batch. Since the generated rules in different batches may contain duplicates, we ultimately employ the LLMs' summarization capability to condense the rules and eliminate duplicates, resulting in the final list of features.

**Interpretable Model Training**

In this stage, all the features identified in the first two stages are transcribed into corresponding functions. All these functions take a scientific instance as input, e.g., a SMILES string for molecules, and return a feature value. Consequently, the final representation of an instance resides in an r-dimensional space, where r is the number of features that have been identified.

These vector representations function as the feature vectors for the model training. Employing interpretable models like a linear layer or random forest enables quantification of each rule's importance in prediction, thus elucidating their contribution to the model's final decision. This transparency fosters an intuitive comprehension of the decision-making process, enhancing trust and usability among domain experts.

**Interpretable Explanation Generation**

The final stage in our pipeline involves generating interpretable explanations for the predictions. Specifically, we furnish the LLMs with salient information, including the model prediction, the vector representation, important rules, and their importance scores derived from the random forest or linear layer. Utilizing the inference and summarization ability of the LLMs, the provided information is transformed into a text-based explanation. This stage is pivotal in rendering the results in an accessible manner. It ensures that users can seamlessly understand the decision-making process and each rule's contribution to the overall prediction, thereby enhancing trust and transparency. This accessibility not only facilitates user interaction with the model but also empowers experts in the field to utilize the generated insights for further analysis and decision-making.

## Metrics

We assessed LLM4SD across 58 molecular property prediction tasks spanning four domains, utilizing distinct evaluation metrics tailored to each task's nature. For the domains of physiology and biophysics, the Area Under the Receiver Operating Characteristic curve (AUC-ROC) metric was employed. AUC-ROC, measures the ability of the model to distinguish between classes, with a range from 0 to 1, where a higher value indicates better performance. In the domain of physical chemistry, the Root Mean Square Error (RMSE) was used. RMSE quantifies the difference between predicted and observed values, with a lower value indicating a closer fit to the true data. For quantum mechanics, we utilized the Mean Absolute Error (MAE) metric. MAE measures the average magnitude of errors between predicted and true values, with smaller values denoting better accuracy.

## Experiment setting

In our experimental setting, we partitioned the data into an 80/10/10 split for training, validation, and test sets. For the domains of physiology, biophysics, and physical chemistry, we employed a scaffold split for molecular compounds. The scaffold split method in these three domains ensures that molecules with similar structures are grouped together, providing a more challenging and realistic evaluation of model generalization. To ensure reproducibility and facilitate further research, the datasets split using this scaffold method are made available in our open-source GitHub repository. In the realm of quantum mechanics, we opted for a random split.

## Data availability

The datasets utilized in this study are entirely open-source and have been made publicly available to ensure straightforward replication of our findings. For research related to quantum mechanics, physical chemistry, biophysics, and physiology, the datasets can be accessed at https://moleculenet.org/datasets-1.

## Code availability

In our commitment to transparency and reproducibility, we will release our code showing our implementation in https://github.com/zyzisastudyreallyhardguy/LLM4SD. This encompasses methodologies for literature knowledge mining, knowledge inference rule mining, interpretable model training, and interpretable explanation generation. Throughout this work, we have employed several open-source libraries, including Hugging Face, numpy, rdkit, pytorch, scipy, bitsandbytes, and accelerate.

Furthermore, we are in the process of deploying a website to facilitate scientists in utilizing LLM4SD. The site features three core functionalities for scientific users: knowledge synthesis, knowledge inference, and prediction with explanations. Examples of user interactions with the website can be found in the supplementary information. As part of our ongoing commitment, we anticipate the inclusion of additional tasks in the future development phases.

38. Park, Yang Jeong, Daniel Kaplan, Zhichu Ren, Chia-Wei Hsu, Changhao Li, Haowei Xu, Sipei Li, and Ju Li. "Can ChatGPT be used to generate scientific hypotheses?." *arXiv preprint arXiv:2304.12208* (2023).

39. Honda, Shion, Shoi Shi, and Hiroki R. Ueda. "Smiles transformer: Pre-trained molecular fingerprint for low data drug discovery." *arXiv preprint arXiv:1911.04738* (2019).

## Acknowledgements:

H.Y.K. scholarship is supported by the Australian Government Research Training Program (RTP) Scholarship and Monash University as a co-contribution to Australian Research Council grant ARC DP210100072. L.T.M, G.W and A.T.N.N research into artificial intelligence applications for drug discovery is supported by a National Health and Medical Research Council (NHMRC) of Australia Ideas grant (APP2013629). Computational resources were generously provided by the Nectar Research Cloud, a collaborative Australian research platform supported by the NCRIS-funded Australian Research Data Commons (ARDC) and the MASSIVE HPC facility. We also gratefully acknowledge the support of the Griffith University eResearch Service & Specialized Platforms Team and the use of the High-Performance Computing Cluster "Gowonda". S.R.P is supported by ARC Future Fellowship (No. FT210100097).

## Author Contribution:

These authors contributed equally: Y.Z.Z., H.Y.K., J.X.J.

These authors jointly supervised this work: S.R.P., G.I.W.

S.R.P. and G.I.W. supervised the project. Y.Z.Z., H.Y.K., J.X.J. contributed to the conception and design of the work. Y.Z.Z., H.Y.K., J.X.J. contributed to the technical implementation. Y.Z.Z., H.Y.K., J.X.J. prepared the figures. Y.Z.Z. contributed to the design of the web-based application. A.T.N.N. and L.T.M. provided domain expertise for the literature review and validation of rules. Y.Z.Z., H.Y.K., A.T.N.N. and L.T.M. contributed to the design of the rule validation test. All authors edited and revised the manuscript.

## Competing Interests:

The authors declare no competing interests.

# Extended Figures and Tables

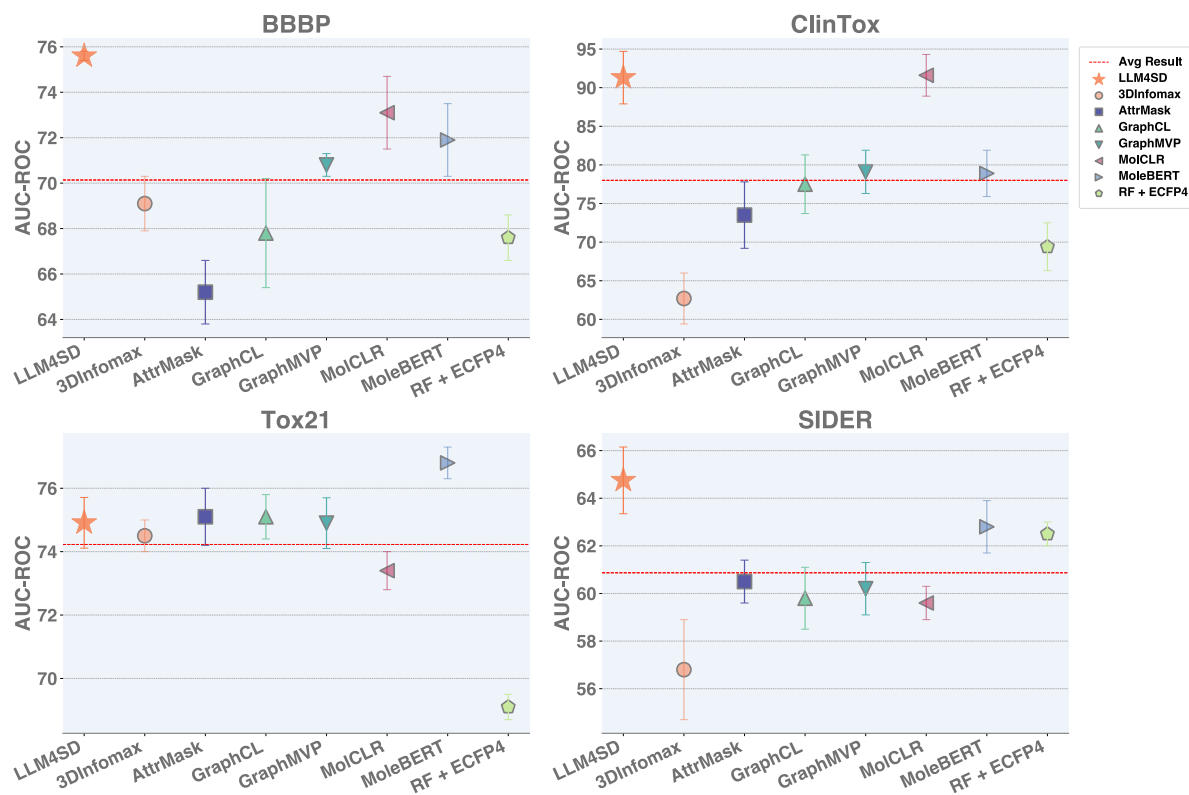

**Extended Data Fig. 1|Detailed performance comparison between "LLM4SD" and eight baselines in the physiology domain.** The red dashed line shows the average result across all methods. Each marker's error bar denotes the method's standard deviation, which is obtained via 10 runs. LLM4SD outperformed other models in 3 out of 4 datasets using the AUC-ROC metric, and consistently surpassing the average across all datasets. The results for Tox21 and SIDER are average scores from 12 and 27 tasks respectively (see Extended Data Fig. 2 and 3 for detailed breakdown).

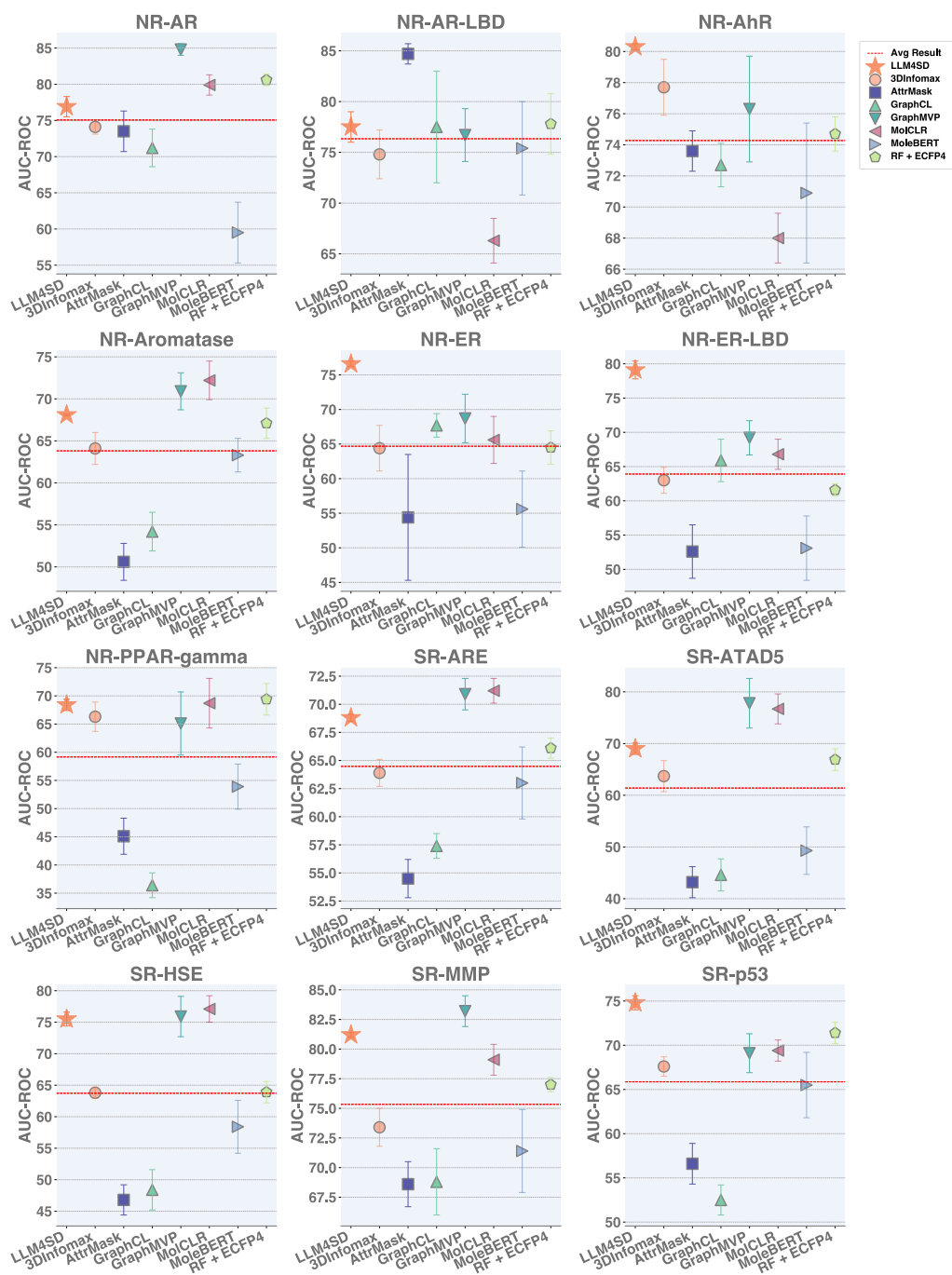

**Extended Data Fig. 2|Detailed performance comparison between "LLM4SD" and eight baselines on Tox21 Dataset.** The red dashed line shows the average result across all methods. Each marker's error bar denotes the method's standard deviation, which is obtained via 10 runs. LLM4SD ranks among the top three methods in 10 out of 12 tasks, and consistently outperformed the average in all tasks.

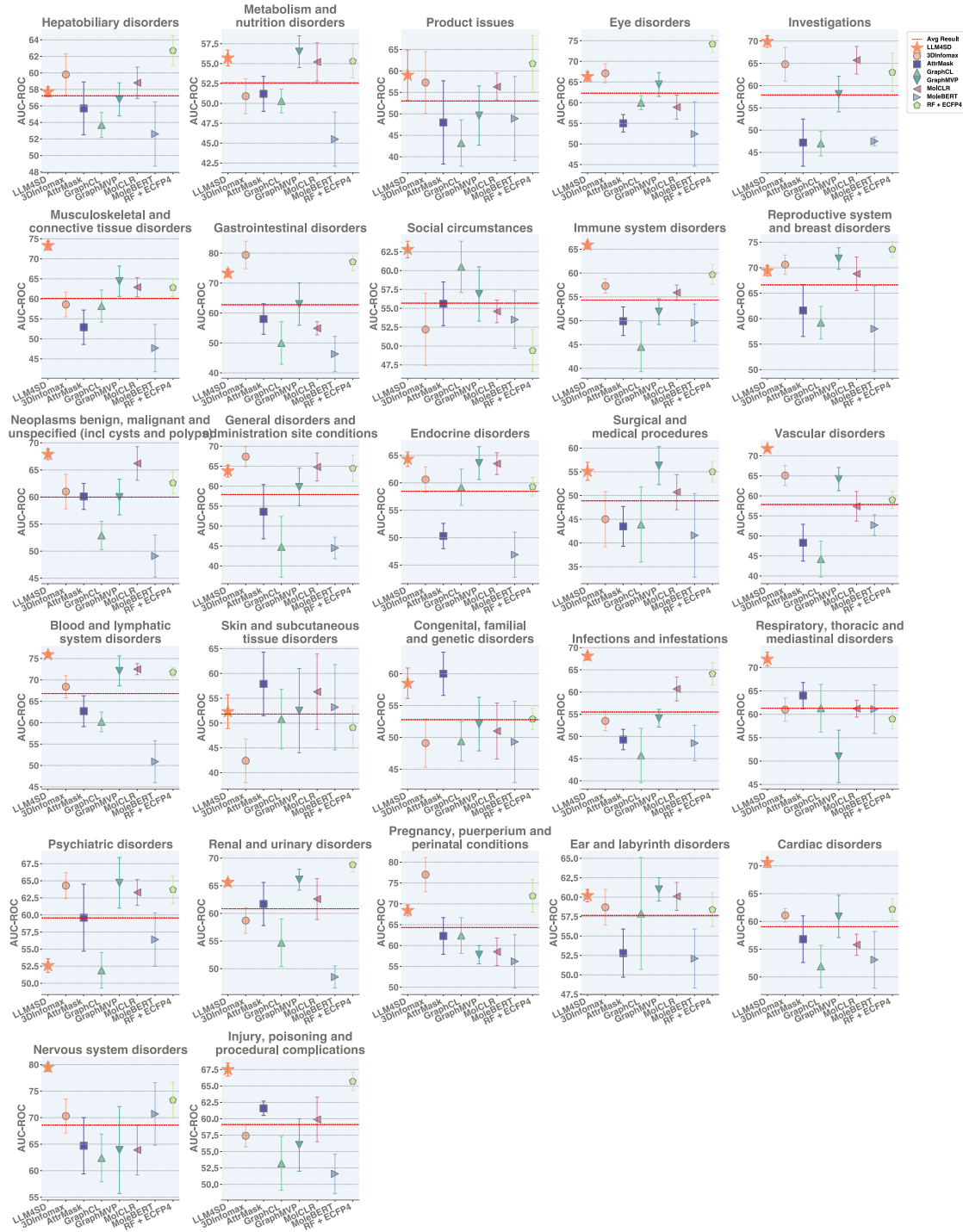

**Extended Data Fig. 3 | Detailed performance comparison between "LLM4SD" and eight baselines on Sider Dataset.** The red dashed line shows the average result across all methods. Each marker's error bar denotes the method's standard deviation, which is obtained via 10 runs. LLM4SD ranks among the top three methods in 22 out of 27 tasks, and consistently outperforms the average in all tasks with the exception of the "Psychiatric disorders" task.

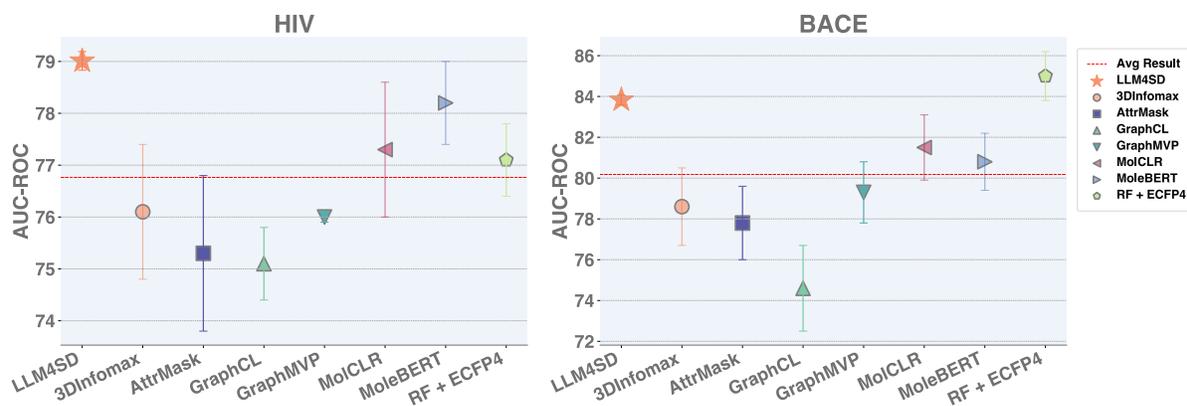

**Extended Data Fig. 4 |Detailed performance comparison between "LLM4SD" and eight baselines in the biophysics domain.** The red dashed line shows the average result across all methods, in terms of AUC-ROC. Each marker's error bar denotes the method's standard deviation, which is obtained via 10 runs. LLM4SD outperformed the top-performing baseline by roughly 2% on the HIV dataset and closely matched the best performing method, RandomForest. In both cases, LLM4SD delivered a visibly superior outcome compared to the average performance.

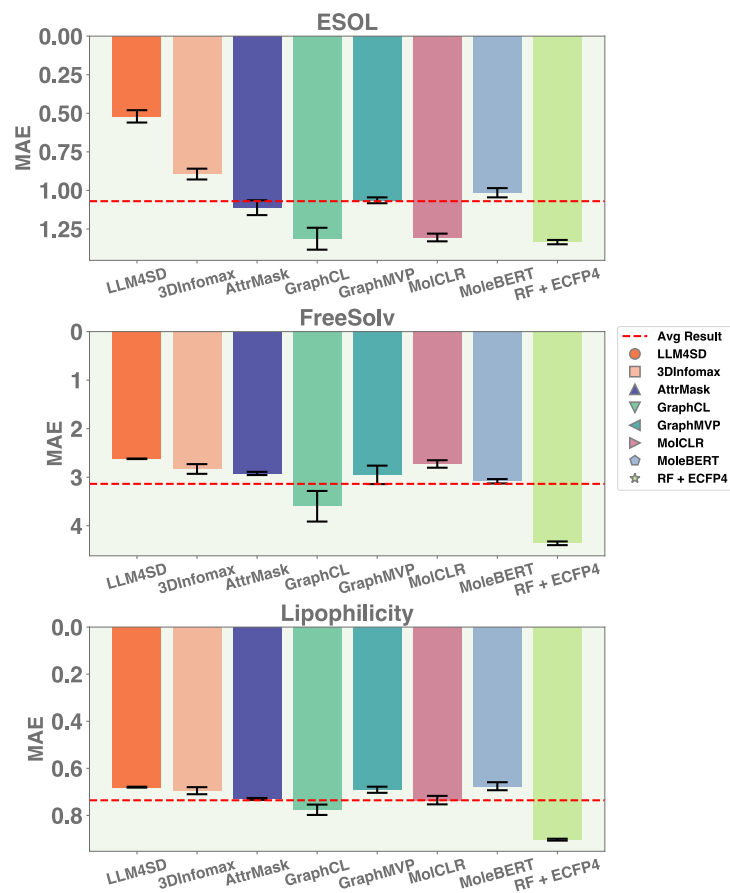

**Extended Data Fig. 5 | Detailed performance comparison between "LLM4SD" and eight baselines in the physical chemistry domain.** The red dashed line shows the average result across all methods. Each marker's error bar denotes the method's standard deviation, which is obtained via 10 runs. LLM4SD significantly outperformed all baseline methods on ESOL, demonstrating a 56% improvement over the average outcome for that dataset, and achieved state-of-the-art results on the additional datasets, FreeSolv and Lipophilicity.

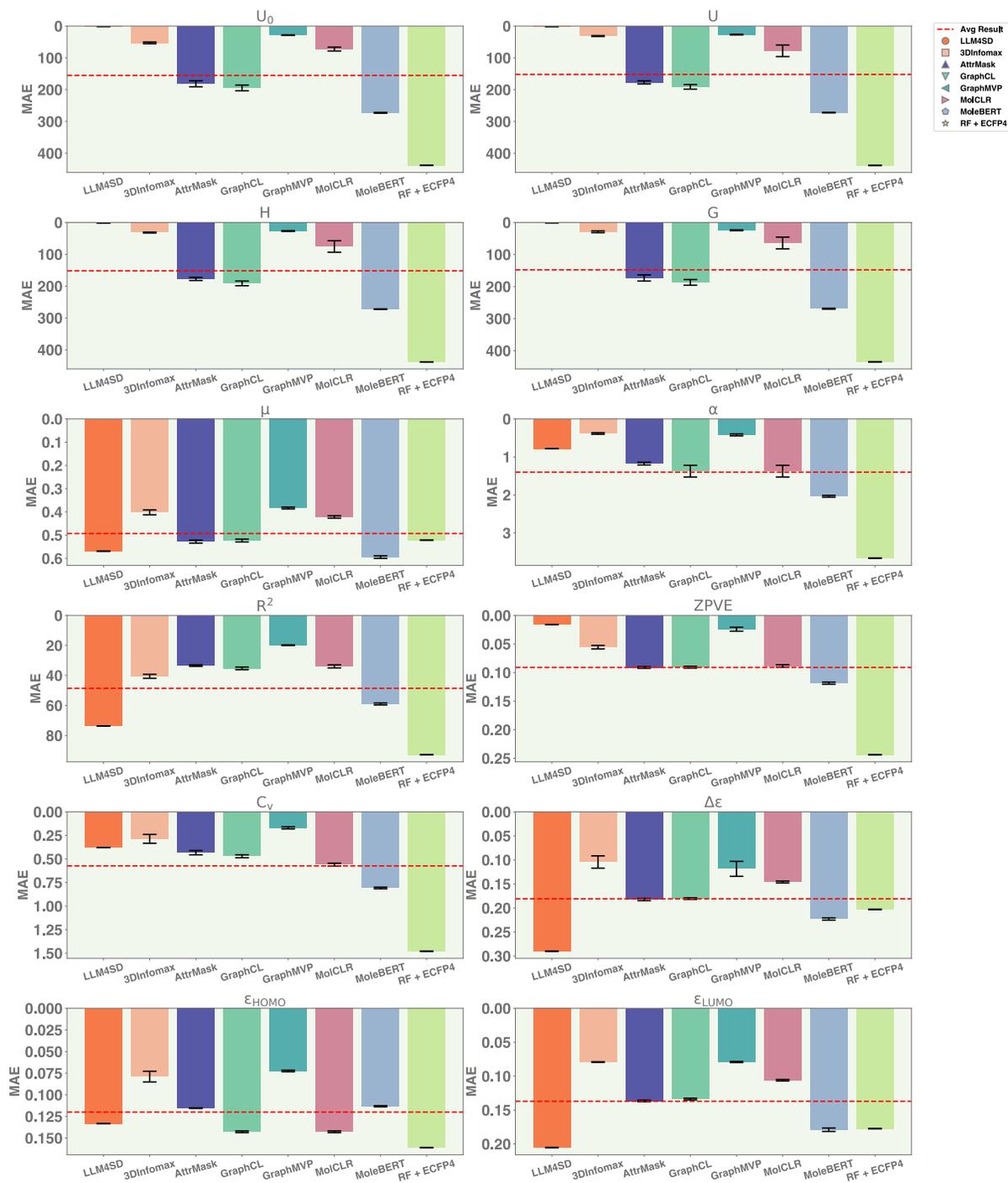

**Extended Data Fig.6| Detailed performance comparison between "LLM4SD" and eight baselines in the quantum mechanics domain.** The red dashed line shows the average result across all methods. Each marker's error bar denotes the method's standard deviation, which is obtained via 10 runs. LLM4SD excelled in predicting properties such as U0, U, H, and G, showing significant enhancements. In other tasks, the results from LLM4SD were comparable to the average of all methods.

**Extended Data Table 1**: General prompt for Knowledge Synthesis from the Scientific Literature and Knowledge Inference from Data.

| |
|---|
| **Prompts for Knowledge Synthesis from the Scientific Literature** |
| **Classification Tasks (General):** <br> Assumed you are an experienced biologist/chemist. Please come up with **[20 rules/30 rules]** that you think are very important to predict **[Task Description]**. Each rule is either about the structure or property of a molecule. <br> **Regression Tasks (General):** <br> Assumed you are an experienced biologist/chemist. Please come up with **[20 rules/30 rules]** that you think are very important to predict **[Task Description]**. Each rule is either about the structure or property of a molecule **(without access to 3D information)**. <br> **Note:** <br> "**20 rules**" is for **smaller LLMs** prompt while "**30 rules**" is for **larger LLMs** prompt. <br> "**without access to 3D information**" is added only for QM9 dataset. |
| **Prompts for Knowledge Inference from Data** |
| **Classification Tasks (General):** <br> Assume you are a very experienced biologist/Chemist. In the following data, with label 1, it means **[Task Description]**. With label 0, it means it is not. Please infer step-by-step to come up with 3 rules that directly relate the properties/structures of a molecule. <br> **Regression Tasks (General):** <br> Assume you are a very experienced biologist/chemist. The following data includes molecules and their corresponding value **[Task Description]**. Please infer step-by-step to come up with 3 rules that directly relate the properties/structures of a molecule **(without access to 3D information)**. <br> **Note:** <br> "**without access to 3D information**" is added only for QM9 dataset. |

**Extended Data Table 2:** Classification task descriptions for the general prompt in Extended Data Table 1.

| Classification Task Name | | | Task Description for Table X General Prompt | | |
|---|---|---|---|---|---|
| **Note: add 'if' for Knowledge Synthesis from the Scientific Literature prompt** | | | | | |
| BBBP | | | **(if)** a molecule is blood brain barrier permeable (BBBP) | | |
| ClinTox | | | **(if)** a molecule will be approved by the FDA | | |
| BACE | | | **(if)** a molecule can inhibit human β-secretase 1(BACE-1) | | |
| HIV | | | **(if)** a molecule can inhibit HIV replication. | | |
| **Note: add 'it is related to' for Knowledge Inference from Data prompt** | | | | | |
| Tox21 | | | **(it is related to)** the toxicity activity of a molecule against the **[subtask description]** in the **[nuclear receptor (NR)/ stress response (SR)]** signalling pathway. | | |
| **Subtask Description** | nr-ar | | androgen receptor | | |
| | nr-ar-lbd | | androgen receptor ligand-binding domain | | |
| | nr-ahr | | aryl hydrocarbon receptor | | |
| | nr-aromatase | | aromatase | | |
| | nr-er | | estrogen receptor | | |
| | nr-er-lbd | | estrogen receptor ligand-binding domain | | |
| | nr-ppar-gamma | | peroxisome proliferator activated receptor | | |
| | sr-are | | nuclear factor (erythroid- derived 2)-like 2 antioxidant responsive element | | |
| | sr-atad5 | | genotoxicity indicated by ATAD5 | | |
| | sr-hse | | heat shock factor response element | | |
| | sr-mmp | | mitochondrial membrane potential | | |
| | sr-p53 | | DNA damage p53-pathway | | |
| Sider | | | **(it is related to)** the side-effect activity of a molecule in causing **[subtask name]**. | | |
| **Subtask names** | respiratory, thoracic and mediastinal disorders | metabolism and nutrition disorders | product issues | eye disorders | investigations |
| | musculoskeletal and connective tissue disorders | blood and lymphatic system disorders | immune system disorders | social circumstances | hepatobiliary disorders |
| | general disorders and administration site conditions | surgical and medical procedures | cardiac disorders | vascular disorders | endocrine disorders |
| | skin and subcutaneous tissue disorders | congenital, familial and genetic disorders | infections and infestations | renal and urinary disorders | psychiatric disorders |
| | pregnancy, puerperium and perinatal conditions | reproductive system and breast disorders | ear and labyrinth disorders | gastrointestinal disorders | nervous system disorders |
| | injury, poisoning and procedural complications | neoplasms benign, malignant and unspecified (incl cysts and polyps) | | | |

**Extended Data Table 3:** Regression task descriptions for the general prompt in Extended Data Table 1.

| Regression Task Name | Task Description for Table X General Prompt |
|---|---|
| ESOL | the water solubility data (log solubility in mols per litre) |
| FreeSolv | the octanol/water distribution coefficient (logD at pH 7.4) |
| Lipophilicity | the hydration free energy of a molecule in water |
| **Quantum Mechanics** | |
| $\mu$ | dipole moment (Mu) of a molecule |
| $\alpha$ | Isotropic polarizability of a molecule |
| $R^2$ | electronic spatial extent of a molecule |
| $ZPVE$ | Zero-Point Vibrational Energy (ZPVE) of a molecule |
| $C_v$ | the heat capacity at constant volume of a molecule |
| $\Delta_\epsilon$ | the HUMO-LUMO gap of a molecule |
| $\epsilon_{HOMO}$ | the highest occupied molecular orbital (HOMO) energy of a molecule |
| $\epsilon_{LUMO}$ | Lowest Unoccupied Molecular Orbital (LUMO) energy of a molecule |
| $U_0$ | internal energy at absolute zero temperature (0 Kelvin), U0, of a molecule |
| U | internal energy of a molecule at a specific temperature, specifically at 298.15 Kelvin (approximately room temperature), (U) of a molecule |
| H | enthalpy of the molecule at a specific temperature, specifically at 298.15 Kelvin (approximately room temperature), (U) of a molecule |
| G | Gibbs free energy of the molecule at a specific temperature, specifically at 298.15 Kelvin (approximately room temperature), (U) of a molecule |